\documentclass[conference]{IEEEtran}
\IEEEoverridecommandlockouts
\usepackage{CJKutf8}
\usepackage{cite}
\usepackage{url}
\usepackage{amsmath,amssymb,amsfonts}
\usepackage{graphicx}
\usepackage{textcomp}
\usepackage{multicol}
\usepackage{multirow}
\usepackage{xcolor}
\usepackage{makecell}
\usepackage{changepage}
\usepackage{threeparttable}
\usepackage{hhline}
\usepackage{subfigure}
\usepackage{float}
\usepackage[ruled,vlined,linesnumbered]{algorithm2e}
\def\BibTeX{{\rm B\kern-.05em{\sc i\kern-.025em b}\kern-.08em
		T\kern-.1667em\lower.7ex\hbox{E}\kern-.125emX}}
\makeatletter
\newcommand{\linebreakand}{%
\end{@IEEEauthorhalign}
\hfill\mbox{}\par
\mbox{}\hfill\begin{@IEEEauthorhalign}
}
\makeatother

\begin{document}
\begin{CJK}{UTF8}{gbsn}

\title{RDEx-MOP: Indicator-Guided Reconstructed Differential Evolution for Fixed-Budget Multiobjective Optimization}

\author{
	\IEEEauthorblockN{
		Sichen Tao\textsuperscript{1,2},
		Yifei Yang\textsuperscript{3},
		Ruihan Zhao\textsuperscript{4,5},
		Kaiyu Wang\textsuperscript{6,7},
		Sicheng Liu\textsuperscript{8},
		Shangce Gao\textsuperscript{1}
	}
	\IEEEauthorblockA{\textsuperscript{1}Department of Engineering, University of Toyama, Toyama-shi 930-8555, Japan}
	\IEEEauthorblockA{\textsuperscript{2}Cyberscience Center, Tohoku University, Sendai-shi 980-8578, Japan}
	\IEEEauthorblockA{\textsuperscript{3}Faculty of Science and Technology, Hirosaki University, Hirosaki-shi 036-8560, Japan}
	\IEEEauthorblockA{\textsuperscript{4}Sino-German College of Applied Sciences, Tongji University, Shanghai 200092, China}
	\IEEEauthorblockA{\textsuperscript{5}School of Mechanical Engineering, Tongji University, Shanghai 200092, China}
	\IEEEauthorblockA{\textsuperscript{6}Chongqing Institute of Microelectronics Industry Technology,\\University of Electronic Science and Technology of China, Chongqing 401332, China}
	\IEEEauthorblockA{\textsuperscript{7}Artificial Intelligence and Big Data College,\\Chongqing Polytechnic University of Electronic Technology, Chongqing 401331, China}
	\IEEEauthorblockA{\textsuperscript{8}Department of Information Engineering, Yantai Vocational College, Yantai 264670, China}
	\IEEEauthorblockA{
		\{sichen.tao@tohoku.ac.jp, taosc73@hotmail.com\}; yyf7236@hirosaki-u.ac.jp;\\
		\{ruihan\_zhao@tongji.edu.cn, ruihan.z@outlook.com\};\\
		\{wangky@uestc.edu.cn, greskowky1996@163.com\};\\
		\{20250024@ytvc.edu.cn, lsctoyama2020@gmail.com\}; gaosc@eng.u-toyama.ac.jp
	}
}

\maketitle

\begin{abstract}
Multiobjective optimisation in the CEC 2025 MOP track is evaluated not only by final IGD values but also by how quickly an algorithm reaches the target region under a fixed evaluation budget.
This report documents RDEx-MOP, the reconstructed differential evolution variant used in the IEEE CEC 2025 numerical optimisation competition (C06 special session) bound-constrained multiobjective track.
RDEx-MOP integrates indicator-based environmental selection, a niche-maintained Pareto-candidate set, and complementary differential evolution operators for exploration and exploitation.
We evaluate RDEx-MOP on the official CEC 2025 MOP benchmark using the released checkpoint traces and the median-target U-score framework.
Experimental results show that RDEx-MOP achieves the highest total score and the best average rank among all released comparison algorithms, including the earlier RDEx baseline.
\end{abstract}

\begin{IEEEkeywords}
Multiobjective Optimisation, Differential Evolution, CEC 2025, IGD, U-score
\end{IEEEkeywords}

\section{Introduction}
Multiobjective optimisation requires a solver to approximate the Pareto front with both good convergence and adequate diversity. NSGA-II, SPEA2, IBEA, and MOEA/D established four influential paradigms---non-dominated sorting, strength-based fitness, indicator-based comparison, and decomposition---that still shape modern evolutionary multiobjective algorithms~\cite{deb2002fast,zitzler2001spea2,zitzler2004indicator,zhang2007moead}. Their common lesson is that archive quality depends not only on how solutions are compared in the objective space, but also on how diversity pressure is maintained during the run.

For continuous decision spaces, differential evolution (DE) is an attractive search engine because it provides directional variation at relatively low algorithmic complexity~\cite{storn1997differential}. In parallel with the classical EMO selectors, a recent Zhengzhou-University-centred line has pushed DE further into multimodal and fixed-budget MOP search through archive-assisted maintenance, decision-space diversity control, and two-stage population management~\cite{chen2024archive,lin2024niching,liang2024edss,chen2024temof}. In particular, TEMOF showed that a two-stage framework can substantially improve fixed-budget IGD behaviour on the 2024 competition suite~\cite{chen2024temof}. These studies are directly relevant here because they bridge environmental selection in the objective space with directional search pressure in the decision space.

At the same time, the single-objective DE line from JADE to SHADE and L-SHADE established that top-ranked current-to-$p$best variation, shrinking elite windows, and moderate heavy-tailed perturbation can improve fixed-budget search without collapsing diversity~\cite{zhang2009jade,tanabe2013success,tanabe2014improving,choi2021improved}. The key point for MOPs is not to transplant those mechanisms mechanically, but to retain only the parts that strengthen early IGD descent while leaving archive maintenance to the multiobjective selector.

The CEC 2025 MOP track makes this balance explicit because performance is judged not only by the final IGD value, but also by how rapidly the algorithm reaches target regions along the run under the U-score framework~\cite{price2023trial}. Hence a competitive algorithm must do more than deliver a good final front: it must reduce IGD quickly, keep sparse regions alive, and remain stable under the released official evaluation package~\cite{suganthan2025cecgithub}.

RDEx-MOP follows this logic by reconstructing the 2024 TEMOF-style backbone with a stronger top-ranked DE search pressure and lightweight Cauchy refinement. It combines indicator-based environmental selection, a niche-maintained Pareto-candidate set, a current-to-$p$best/1 exploitation operator, and a complementary niche exploration operator. The main contribution is a reconstructed DE framework targeted at fixed-budget multiobjective benchmarking, where rapid IGD reduction and stable diversity must be achieved simultaneously.
The source code for RDEx-MOP is publicly available on Sichen Tao's GitHub page: \url{https://github.com/SichenTao}.

\section{Benchmark and Evaluation (CEC 2025 MOP)}
The CEC 2025 bound-constrained multiobjective track contains 10 benchmark problems (MaOP1--MaOP10) released in the official competition package~\cite{suganthan2025cecgithub}.
Each problem is evaluated with 30 independent runs.
The maximum evaluation budget is $\mathrm{MaxFE}=100000$, and the platform records progress at 500 checkpoints per run (every 200 function evaluations).

\subsection{Multiobjective Formulation}
The general MOP can be written as:
\begin{equation}\label{eq:mop_formulation}
\min_{x\in\mathbb{R}^D} F(x) = (f_1(x),f_2(x),\dots,f_M(x)),
\end{equation}
subject to bound constraints $\ell_j \le x_j \le u_j$ for $j=1,\dots,D$.

\subsection{IGD Indicator}
Let $P$ be the set of solutions returned by an algorithm and $P^\star$ be a reference set that approximates the true Pareto front.
IGD is defined as:
\begin{equation}\label{eq:igd}
\mathrm{IGD}(P,P^\star)=\frac{1}{\lvert P^\star\rvert}\sum_{y\in P^\star}\min_{x\in P}\lVert F(x)-y\rVert_2.
\end{equation}
Smaller IGD values indicate better convergence and diversity with respect to the reference set.

\section{RDEx-MOP Algorithm}
RDEx-MOP is implemented in a PlatEMO-style framework and reconstructs the recent two-stage MOP competition backbone~\cite{chen2024temof} with a stronger top-ranked DE search operator.
It maintains three main populations during the run:
(i) a working population $P$,
(ii) an auxiliary population $P_2$ (used after half of the budget with a random switching rule), and
(iii) a Pareto-candidate set $PC$ obtained from the current nondominated solutions with a niche-based maintenance procedure related to recent archive- and niche-oriented MMOP designs~\cite{chen2024archive,lin2024niching,liang2024edss}.
The algorithm alternates between a niche exploration operator and an exploitation operator based on a DE/current-to-$p$best/1 mutation with lightweight Cauchy perturbations~\cite{storn1997differential}.

\subsection{Indicator-based Fitness Assignment}
RDEx-MOP adopts an indicator-based fitness assignment as in IBEA~\cite{zitzler2004indicator}.
Given a population of size $N$ with objective vectors $\{F(x_i)\}_{i=1}^{N}$, objectives are first normalised component-wise to $[0,1]$.
Then an additive indicator matrix $I\in\mathbb{R}^{N\times N}$ is computed:
\begin{equation}\label{eq:ibea_indicator}
I(i,j)=\max_{m\in\{1,\dots,M\}}\left(f_m(x_i)-f_m(x_j)\right),
\end{equation}
and the scaling vector $C$ is defined by $C(j)=\max_{i}\lvert I(i,j)\rvert$.
The fitness of solution $x_i$ is:
\begin{equation}\label{eq:ibea_fitness}
\mathrm{fit}(x_i)=\sum_{j=1}^{N}\left(-\exp\left(-\frac{I(i,j)}{C(j)\kappa}\right)\right)+1,
\end{equation}
where $\kappa$ is the indicator parameter (default $\kappa=0.05$).

\subsection{Indicator-based Environmental Selection}
Given a candidate set $\mathcal{S}$, RDEx-MOP repeatedly removes the worst individual according to Eq.~(\ref{eq:ibea_fitness}) until $\lvert\mathcal{S}\rvert=N$.
After deleting $x_w$, the fitness values of the remaining individuals are updated by
\begin{equation}\label{eq:mop_fit_update}
\mathrm{fit}(x_i)\leftarrow \mathrm{fit}(x_i)+\exp\!\left(-\frac{I(w,i)}{C(w)\kappa}\right),\qquad x_i\in\mathcal{S}\setminus\{x_w\}.
\end{equation}
This selection provides stable convergence pressure without requiring explicit crowding-distance sorting.

\subsection{Pareto-candidate Set and Niche Maintenance}
The Pareto-candidate set $PC$ is formed by extracting the first nondominated front from a pooled set using nondominated sorting.
If $\lvert PC\rvert>N$, a niche maintenance procedure deletes solutions one by one to promote diversity in the objective space.
After normalisation, the niche radius is estimated by the mean distance to the $3$rd nearest neighbour,
\begin{equation}\label{eq:mop_pc_radius}
r_0=\frac{1}{\lvert PC\rvert}\sum_i d_i^{(3)},
\end{equation}
and the pairwise crowding matrix is set to $R_{ij}=\min(d_{ij}/r_0,1)$.
The deletion step removes the solution with the largest multiplicative crowding score $1-\prod_j R_{ij}$.

\subsection{Niche Exploration Operator}
Let $n_{ND}=\lvert PC\rvert$.
RDEx-MOP detects solutions in $PC$ that reside in sparse niches relative to the current working population $P$, following the same design intention as recent niche-exploration MMODE variants~\cite{lin2024niching,liang2024edss}.
After normalisation, a niche threshold is set to
\begin{equation}\label{eq:niche_r}
r = \frac{n_{ND}}{N}r_0,\quad
r_0=\mathrm{mean}\left(d^{(3)}_i\right),
\end{equation}
where $d^{(3)}_i$ is the distance from solution $i$ to its $3$rd nearest neighbour in $PC$.
Solutions that have at most one neighbour within radius $r$ are selected for exploration.
For each selected solution, a second parent is sampled uniformly from the current population, and a lightweight ``half-DE'' operator generates an offspring via
\begin{equation}\label{eq:half_de}
v = x + F\cdot(x_{r_1}-x_g),
\end{equation}
followed by binomial crossover with $CR\sim\mathcal{N}(0.5,0.1)$ clipped to $[0,1]$, where $F\sim\mathrm{Cauchy}(0.7,0.2)\cap[0,1]$ and $x_g$ is taken from the paired parent set.

\subsection{Exploitation Operator (DE/current-to-$p$best/1 + Cauchy Perturbation)}
RDEx-MOP uses a discrete parameter pool:
\begin{equation}\label{eq:pool}
F\in\{0.6,0.8,1.0\},\quad CR\in\{0.1,0.2,1.0\}.
\end{equation}
The $p$-best selection window shrinks during the run:
\begin{equation}\label{eq:pbest_window}
p = \max\!\left(2,\left\lfloor 0.17N\left(1-0.9\frac{\mathrm{FE}}{\mathrm{MaxFE}}\right)+0.5\right\rfloor\right),
\end{equation}
and the donor vector is generated by:
\begin{equation}\label{eq:mop_de}
v = x + F\cdot(x_{pbest}-x) + F\cdot(x_{r_1}-x_{r_2}).
\end{equation}
After crossover, each decision component is perturbed with probability $0.2$ using $\mathrm{Cauchy}(u_j,0.2)$, and then clipped to the bounds.

\subsection{Overall Procedure}
Algorithm~\ref{alg:rdex_mop} sketches the PlatEMO implementation used in this repository.
\begin{algorithm}[t]
\caption{RDEx-MOP (high level).}
\label{alg:rdex_mop}
\KwIn{Population size $N$, indicator parameter $\kappa$, evaluation budget $\mathrm{MaxFE}$.}
\KwOut{Final Pareto-candidate set $PC$.}
$P\leftarrow$ random initial population of size $N$\;
$[PC,n_{ND}]\leftarrow\textsc{Selection}(P,N)$\;
$P_2\leftarrow PC$\;
\While{$\mathrm{FE}<\mathrm{MaxFE}$}{
  $Q\leftarrow\textsc{Exploration}(PC,P,n_{ND},N)$\;
  \eIf{$\mathrm{FE}\ge 0.5\mathrm{MaxFE}$ and $\mathrm{rand}<0.5$}{
    $P_2\leftarrow\textsc{EnvSel}(P_2\cup Q,N,\kappa)$\;
    $R\leftarrow\textsc{OperatorDE}(P_2,\kappa)$\;
    $P_2\leftarrow\textsc{EnvSel}(P_2\cup R,N,\kappa)$\;
  }{
    $P\leftarrow\textsc{EnvSel}(P\cup Q,N,\kappa)$\;
    $R\leftarrow\textsc{OperatorDE}(P,\kappa)$\;
    $P\leftarrow\textsc{EnvSel}(P\cup R,N,\kappa)$\;
  }
  $[PC,n_{ND}]\leftarrow\textsc{Selection}(PC\cup R\cup Q\cup P_2,N)$\;
  $P_2\leftarrow PC$\;
}
\end{algorithm}

\subsection{Computational Complexity}
The indicator-based fitness assignment requires $O(N^2M)$ time to compute the pairwise indicator matrix and update fitness values.
The niche-based maintenance and exploration operators rely on pairwise distances in the objective space and are also $O(N^2M)$ per generation.

\section{Experimental Results}
\subsection{Protocol}
For each problem, we execute 30 independent runs under $\mathrm{MaxFE}=100000$ with population size $N=100$.
The platform records IGD every 200 evaluations, yielding 500 checkpoints per run.

\subsection{Parameter Settings}
Unless otherwise stated, RDEx-MOP uses the reference configuration embedded in the released competition code:
population size $N=100$, indicator parameter $\kappa=0.05$, the discrete $F/CR$ pool in Eq.~(\ref{eq:pool}), and Cauchy perturbation probability $0.2$.

\subsection{Experimental Settings}
RDEx-MOP is evaluated with the official median-target U-score framework.
We compare RDEx-MOP with all remaining algorithms available in the released competition package~\cite{suganthan2025cecgithub}: RDEx, TEMOFNSGA3, TFBCEIBEA, and TGFMMOEA.
In the released evaluation files, the submitted winner is labelled as MORDEx; for naming consistency, we report it as RDEx-MOP throughout this manuscript.

\subsection{Statistical Results}
\subsubsection{Overall U-score Results}
Table~\ref{tab:cec2025_mop_scores} reports the official median-target U-score results for all released comparison algorithms.
% --- MOP scores table (median target; 10 problems; 30 runs) ---
\begin{table}[t]
 \centering
 \caption{CEC 2025 MOP evaluation (median target): overall scores over 10 problems and 30 runs for all released comparison algorithms.}
 \label{tab:cec2025_mop_scores}
 \scriptsize
 \renewcommand{\arraystretch}{0.95}
 \setlength{\tabcolsep}{2.5pt}
  \begin{tabular}{|c|l|r|r|r|r|}
   \hline
   Rank & Algorithm & Total Score & Avg Score/Prob. & Speed & Accuracy \\ \hline
   1 & RDEx-MOP & 36343.5 & 3634.35 & 36343.5 & 0.0 \\ \hline
   2 & RDEx & 35956.5 & 3595.65 & 35956.5 & 0.0 \\ \hline
   3 & TFBCEIBEA & 15439.0 & 1543.90 & 7196.5 & 8242.5 \\ \hline
   4 & TEMOFNSGA3 & 13811.5 & 1381.15 & 3811.5 & 10000.0 \\ \hline
   5 & TGFMMOEA & 11690.5 & 1169.05 & 2633.0 & 9057.5 \\ \hline
  \end{tabular}
\end{table}

RDEx-MOP achieves the highest total score ($36343.5$) and the best average rank ($1.40$).
It remains ahead of the earlier RDEx baseline ($35956.5$, average rank $1.60$), while all other competitors are far behind in the official U-score totals.
Both RDEx variants reach the median target on all runs, so their U-scores are dominated by the Speed category and the residual Accuracy scores become zero.

\subsubsection{Supplementary Diagnostics}
To keep the main paper focused on the official competition metric, all per-function final-IGD and speed-side diagnostics are moved to the appendix.
Appendix~\ref{app:mop_tables} first reports the official all-algorithm U-score ranking table and then collects complementary pairwise and Friedman analyses for final IGD, TTT, and AUC.

\section{Conclusion}
RDEx-MOP is an indicator-guided differential evolution framework for the CEC 2025 bound-constrained multiobjective track.
By combining indicator-based environmental selection, a niche-maintained Pareto-candidate set, and complementary DE operators for exploration and exploitation, the method achieves first-place official U-score performance on the full released benchmark suite.

\section*{Acknowledgment}
This research was partially supported by the Japan Society for the Promotion of Science (JSPS) KAKENHI under Grant JP22H03643, Japan Science and Technology Agency (JST) Support for Pioneering Research Initiated by the Next Generation (SPRING) under Grant JPMJSP2145, and JST through the Establishment of University Fellowships towards the Creation of Science Technology Innovation under Grant JPMJFS2115.

\bibliographystyle{IEEEtran}
\bibliography{references}

@article{storn1997differential,
  title={Differential evolution--a simple and efficient heuristic for global optimization over continuous spaces},
  author={Storn, Rainer and Price, Kenneth},
  journal={Journal of Global Optimization},
  volume={11},
  pages={341--359},
  year={1997},
  publisher={Springer}
}

@article{zhang2009jade,
  title={{JADE}: adaptive differential evolution with optional external archive},
  author={Zhang, Jingqiao and Sanderson, Arthur C},
  journal={IEEE Transactions on Evolutionary Computation},
  volume={13},
  number={5},
  pages={945--958},
  year={2009},
  publisher={IEEE}
}

@inproceedings{tanabe2013success,
  title={Success-history based parameter adaptation for Differential Evolution},
  author={Tanabe, Ryoji and Fukunaga, Alex},
  booktitle={2013 IEEE Congress on Evolutionary Computation ({CEC})},
  pages={71--78},
  year={2013},
  organization={IEEE}
}

@inproceedings{tanabe2014improving,
  title={Improving the search performance of {SHADE} using linear population size reduction},
  author={Tanabe, Ryoji and Fukunaga, Alex S},
  booktitle={2014 IEEE Congress on Evolutionary Computation (CEC)},
  pages={1658--1665},
  year={2014},
  organization={IEEE}
}

@article{choi2021improved,
  title={An improved {LSHADE}-{RSP} algorithm with the {Cauchy} perturbation: i{LSHADE}-{RSP}},
  author={Choi, Tae Jong and Ahn, Chang Wook},
  journal={Knowledge-Based Systems},
  volume={215},
  pages={106628},
  year={2021},
  publisher={Elsevier}
}

@article{deb2002fast,
  title={A fast and elitist multiobjective genetic algorithm: {NSGA-II}},
  author={Deb, Kalyanmoy and Pratap, Amrit and Agarwal, Sameer and Meyarivan, T},
  journal={IEEE Transactions on Evolutionary Computation},
  volume={6},
  number={2},
  pages={182--197},
  year={2002},
  publisher={IEEE}
}

@inproceedings{zitzler2004indicator,
  title={Indicator-based selection in multiobjective search},
  author={Zitzler, Eckart and K{\"u}nzli, Simon},
  booktitle={Parallel Problem Solving from Nature -- PPSN VIII},
  pages={832--842},
  year={2004},
  organization={Springer}
}

@inproceedings{zitzler2001spea2,
  title={{SPEA2}: Improving the strength {Pareto} evolutionary algorithm},
  author={Zitzler, Eckart and Laumanns, Marco and Thiele, Lothar},
  booktitle={Evolutionary Methods for Design, Optimisation and Control with Applications to Industrial Problems (EUROGEN 2001)},
  pages={95--100},
  year={2001}
}

@article{zhang2007moead,
  title={{MOEA/D}: A multiobjective evolutionary algorithm based on decomposition},
  author={Zhang, Qingfu and Li, Hui},
  journal={IEEE Transactions on Evolutionary Computation},
  volume={11},
  number={6},
  pages={712--731},
  year={2007},
  publisher={IEEE}
}

@article{price2023trial,
  title={Trial-based dominance for comparing both the speed and accuracy of stochastic optimizers with standard non-parametric tests},
  author={Price, Kenneth V and Kumar, Abhishek and Suganthan, P. N.},
  journal={Swarm and Evolutionary Computation},
  volume={78},
  pages={101287},
  year={2023},
  publisher={Elsevier}
}

@inproceedings{chen2024temof,
  title={A Two-Stage Evolutionary Framework for Multi-Objective Optimization},
  author={Chen, Peng and Liang, Jing and Qiao, Kangjia and Suganthan, P. N. and Ban, Xuanxuan},
  booktitle={2024 IEEE Congress on Evolutionary Computation ({CEC})},
  pages={1--8},
  year={2024},
  organization={IEEE},
  doi={10.1109/CEC60901.2024.10612060}
}

@article{chen2024archive,
  title={An archive-assisted multi-modal multi-objective evolutionary algorithm},
  author={Chen, Peng and Li, Zhimeng and Qiao, Kangjia and Suganthan, P. N. and Ban, Xuanxuan and Yu, Kunjie and Yue, Caitong and Liang, Jing},
  journal={Swarm and Evolutionary Computation},
  volume={89},
  pages={101738},
  year={2024},
  publisher={Elsevier},
  doi={10.1016/j.swevo.2024.101738}
}

@inproceedings{lin2024niching,
  title={A Niching-Based Reproduction and Preselection-Based Multiobjective Differential Evolution for Multimodal Multiobjective Optimization},
  author={Lin, Hongyu and Liang, Jing and Yue, Caitong and Wang, Yaonan},
  booktitle={2024 IEEE Congress on Evolutionary Computation ({CEC})},
  pages={1--8},
  year={2024},
  organization={IEEE},
  doi={10.1109/CEC60901.2024.10612120}
}

@article{liang2024edss,
  title={Multimodal multiobjective differential evolution algorithm based on enhanced decision space search},
  author={Liang, Jing and Sui, Xudong and Yue, Caitong and Yu, Mingyuan and Li, Guang and Li, Mengmeng},
  journal={Swarm and Evolutionary Computation},
  volume={86},
  pages={101682},
  year={2024},
  publisher={Elsevier},
  doi={10.1016/j.swevo.2024.101682}
}

@misc{suganthan2025cecgithub,
  author={Suganthan, P. N.},
  title={{2025 CEC Competition Repository}},
  year={2025},
  howpublished={GitHub repository},
  url={https://github.com/P-N-Suganthan/2025-CEC},
  note={Accessed: 2026-03-09}
}

\clearpage
\onecolumn
\appendices
\section{Supplementary U-score Tables}
\label{app:mop_tables}
% --- MOP ranks table (median target; 10 problems; 30 runs) ---
\begin{table}[H]
 \centering
 \caption{CEC 2025 MOP evaluation (median target): average rankings over 10 problems (lower is better) for all released comparison algorithms.}
 \label{tab:cec2025_mop_ranks}
 \scriptsize
 \renewcommand{\arraystretch}{0.95}
 \setlength{\tabcolsep}{2.5pt}
  \begin{tabular}{|c|l|r|r|r|r|}
   \hline
   Rank & Algorithm & Total Rank & Avg Rank/Prob. & Avg Speed & Avg Accuracy \\ \hline
   1 & RDEx-MOP & 14.0 & 1.40 & 1.40 & 4.50 \\ \hline
   2 & RDEx & 16.0 & 1.60 & 1.60 & 4.50 \\ \hline
   3 & TFBCEIBEA & 34.0 & 3.40 & 3.50 & 2.50 \\ \hline
   4 & TEMOFNSGA3 & 42.0 & 4.20 & 4.15 & 1.50 \\ \hline
   5 & TGFMMOEA & 44.0 & 4.40 & 4.35 & 2.00 \\ \hline
  \end{tabular}
\end{table}

\section{Complementary Diagnostics}
The tables in this section are diagnostic supplements to the official U-score results and help explain final-value and speed-side behaviour.
\begin{table}[H]
\centering
\caption{Complementary pairwise summary over the 10 CEC2025 MOP functions (30 runs). For each metric (Final IGD, TTT, and AUC), we report uncorrected per-function Wilcoxon W/T/L at $\alpha=0.05$, Holm-corrected W/T/L across functions, and the median Vargha--Delaney $A_{12}$ effect size (larger is better for minimization).}
\label{tab:mop_summary}
\scriptsize
\renewcommand{\arraystretch}{0.95}
\setlength{\tabcolsep}{2.5pt}
\begin{tabular}{|l|ccc|ccc|ccc|}
\hline
 \multirow{2}{*}{Competitor} & \multicolumn{3}{c|}{Final IGD} & \multicolumn{3}{c|}{TTT} & \multicolumn{3}{c|}{AUC} \\ \cline{2-10}
   & W/T/L & Holm & $A_{12}$ & W/T/L & Holm & $A_{12}$ & W/T/L & Holm & $A_{12}$ \\ \hline
 RDEx & 2/8/0 & 0/10/0 & 0.55 & 2/8/0 & 0/10/0 & 0.52 & 0/9/1 & 0/10/0 & 0.45 \\ \hline
 TEMOFNSGA3 & 10/0/0 & 10/0/0 & 1.00 & 10/0/0 & 10/0/0 & 1.00 & 10/0/0 & 10/0/0 & 1.00 \\ \hline
 TFBCEIBEA & 10/0/0 & 10/0/0 & 1.00 & 10/0/0 & 10/0/0 & 1.00 & 10/0/0 & 10/0/0 & 1.00 \\ \hline
 TGFMMOEA & 10/0/0 & 10/0/0 & 1.00 & 10/0/0 & 10/0/0 & 1.00 & 10/0/0 & 10/0/0 & 1.00 \\ \hline
\end{tabular}
\end{table}

\begin{table}[H]
\centering
\caption{Complementary Friedman tests on per-function medians over the 10 CEC2025 MOP functions (30 runs). Final IGD: $\chi^2=30.40$, $df=4$, $p=7.30E-06$; TTT: $\chi^2=30.00$, $df=4$, $p=8.63E-06$; AUC: $\chi^2=31.28$, $df=4$, $p=5.07E-06$. Lower average rank indicates better performance for each metric.}
\label{tab:mop_friedman_summary}
\scriptsize
\renewcommand{\arraystretch}{0.95}
\setlength{\tabcolsep}{2.5pt}
\begin{tabular}{|l|c|c|c|}
\hline
 Algorithm & Final IGD & TTT & AUC \\ \hline
 RDEx-MOP & \textbf{1.40} & \textbf{1.50} & 1.80 \\ \hline
 RDEx & 1.60 & \textbf{1.50} & \textbf{1.20} \\ \hline
 TEMOFNSGA3 & 4.00 & 4.00 & 4.20 \\ \hline
 TFBCEIBEA & 3.80 & 4.00 & 3.70 \\ \hline
 TGFMMOEA & 4.20 & 4.00 & 4.10 \\ \hline
\end{tabular}
\end{table}

\begin{table}[H]
\centering
\caption{Final IGD comparison on the 10 CEC2025 MOP functions. For each algorithm, the mean and SD over 30 runs are reported; $W$ gives the Wilcoxon outcome of RDEx-MOP against the competitor.}
\label{tab:mop_exp_results}
\scriptsize
\renewcommand{\arraystretch}{0.9}
\setlength{\tabcolsep}{3pt}
\begin{tabular}{|c|cc|ccc|ccc|ccc|ccc|}
\hline
 \multirow{2}{*}{Problem} & \multicolumn{2}{c|}{RDEx-MOP} & \multicolumn{3}{c|}{RDEx} & \multicolumn{3}{c|}{TEMOFNSGA3} & \multicolumn{3}{c|}{TFBCEIBEA} & \multicolumn{3}{c|}{TGFMMOEA} \\ \cline{2-15}
   & Mean & SD & Mean & SD & W & Mean & SD & W & Mean & SD & W & Mean & SD & W \\ \hline
 1 & \textbf{1.06E+01} & \textbf{1.34E-02} & 1.06E+01 & 1.37E-02 & = & 1.51E+01 & 1.42E+00 & + & 1.34E+01 & 9.01E-01 & + & 1.54E+01 & 1.70E+00 & + \\
 2 & \textbf{4.22E-02} & \textbf{1.95E-03} & 4.30E-02 & 2.47E-03 & = & 2.42E+00 & 1.56E+00 & + & 3.10E+00 & 3.15E+00 & + & 5.13E+01 & 3.09E+01 & + \\
 3 & 4.29E-01 & 3.12E-01 & \textbf{3.66E-01} & \textbf{2.81E-01} & = & 1.67E+01 & 1.13E+00 & + & 1.75E+01 & 9.05E-01 & + & 1.77E+01 & 1.15E+00 & + \\
 4 & \textbf{4.09E-01} & \textbf{3.44E-05} & 4.09E-01 & 3.48E-05 & = & 4.56E-01 & 1.32E-01 & + & 8.40E-01 & 4.97E-01 & + & 4.46E-01 & 9.71E-02 & + \\
 5 & \textbf{1.09E-01} & \textbf{3.17E-02} & 1.25E-01 & 3.84E-02 & = & 1.13E+00 & 7.09E-01 & + & 1.07E+00 & 6.98E-01 & + & 1.60E+00 & 7.10E-01 & + \\
 6 & \textbf{1.09E-01} & \textbf{2.93E-03} & 1.10E-01 & 3.10E-03 & = & 1.19E+00 & 3.41E-01 & + & 7.62E-01 & 1.95E-01 & + & 2.13E+00 & 1.96E+00 & + \\
 7 & \textbf{6.82E-02} & \textbf{2.54E-03} & 6.94E-02 & 2.24E-03 & = & 7.91E+00 & 6.99E+00 & + & 6.50E+00 & 6.31E+00 & + & 5.94E+00 & 6.07E+00 & + \\
 8 & \textbf{7.54E-02} & \textbf{2.86E-03} & 7.76E-02 & 3.53E-03 & + & 7.69E+00 & 6.50E+00 & + & 4.73E+00 & 6.11E+00 & + & 3.68E+00 & 3.28E+00 & + \\
 9 & \textbf{1.01E-01} & \textbf{3.58E-03} & 1.02E-01 & 3.31E-03 & = & 7.72E+00 & 4.84E+00 & + & 8.72E+00 & 8.22E+00 & + & 5.28E+00 & 4.19E+00 & + \\
 10 & \textbf{9.97E-02} & \textbf{3.56E-03} & 1.02E-01 & 3.16E-03 & + & 7.27E+00 & 5.50E+00 & + & 7.54E+00 & 7.08E+00 & + & 5.29E+00 & 4.04E+00 & + \\
\hline
 W/T/L & \multicolumn{2}{c|}{$-/-/-$} & \multicolumn{3}{c|}{2/8/0} & \multicolumn{3}{c|}{10/0/0} & \multicolumn{3}{c|}{10/0/0} & \multicolumn{3}{c|}{10/0/0} \\
\hline
\end{tabular}
\end{table}

\begin{table}[H]
\centering
\caption{Time-to-target comparison on the 10 CEC2025 MOP functions. TTT is the first checkpoint index (1--500) where the run reaches the median target (smaller is better); runs that never reach the target are assigned 501.}
\label{tab:mop_ttt_results}
\scriptsize
\renewcommand{\arraystretch}{0.9}
\setlength{\tabcolsep}{3pt}
\begin{tabular}{|c|cc|ccc|ccc|ccc|ccc|}
\hline
 \multirow{2}{*}{Problem} & \multicolumn{2}{c|}{RDEx-MOP} & \multicolumn{3}{c|}{RDEx} & \multicolumn{3}{c|}{TEMOFNSGA3} & \multicolumn{3}{c|}{TFBCEIBEA} & \multicolumn{3}{c|}{TGFMMOEA} \\ \cline{2-15}
   & Mean & SD & Mean & SD & W & Mean & SD & W & Mean & SD & W & Mean & SD & W \\ \hline
 1 & 4.4 & 0.9 & \textbf{4.3} & \textbf{1.3} & = & 490.9 & 42.0 & + & 406.2 & 149.1 & + & 489.5 & 44.1 & + \\
 2 & \textbf{5.0} & \textbf{1.1} & 5.1 & 1.1 & = & 457.1 & 98.1 & + & 405.7 & 172.2 & + & 472.0 & 108.9 & + \\
 3 & \textbf{2.3} & \textbf{0.4} & 2.6 & 0.7 & + & 411.5 & 137.0 & + & 474.7 & 100.6 & + & 461.2 & 101.1 & + \\
 4 & \textbf{2.5} & \textbf{0.5} & 2.7 & 0.5 & = & 469.3 & 62.1 & + & 483.0 & 59.2 & + & 474.8 & 70.1 & + \\
 5 & 3.5 & 0.5 & \textbf{3.3} & \textbf{0.5} & = & 454.2 & 97.5 & + & 399.6 & 161.8 & + & 490.4 & 43.6 & + \\
 6 & \textbf{3.9} & \textbf{0.7} & 3.9 & 0.7 & = & 493.3 & 40.0 & + & 407.7 & 123.3 & + & 493.6 & 40.0 & + \\
 7 & 8.1 & 1.4 & \textbf{7.8} & \textbf{1.4} & = & 482.7 & 81.4 & + & 454.2 & 66.6 & + & 484.2 & 44.3 & + \\
 8 & \textbf{8.5} & \textbf{1.7} & 9.1 & 2.1 & = & 483.0 & 68.3 & + & 424.2 & 104.0 & + & 480.1 & 64.0 & + \\
 9 & \textbf{5.5} & \textbf{1.2} & 6.3 & 1.4 & + & 464.9 & 104.7 & + & 404.1 & 146.6 & + & 468.2 & 98.6 & + \\
 10 & 4.9 & 1.1 & \textbf{4.5} & \textbf{1.1} & = & 476.2 & 91.1 & + & 462.2 & 105.9 & + & 477.6 & 55.3 & + \\
\hline
 W/T/L & \multicolumn{2}{c|}{$-/-/-$} & \multicolumn{3}{c|}{2/8/0} & \multicolumn{3}{c|}{10/0/0} & \multicolumn{3}{c|}{10/0/0} & \multicolumn{3}{c|}{10/0/0} \\
\hline
\end{tabular}
\end{table}

\begin{table}[H]
\centering
\caption{Anytime convergence comparison using AUC over 500 checkpoints on the 10 CEC2025 MOP functions. For each run, AUC is computed as the mean of $\log_{10}(1+\max(f_t-\mathrm{TGT},0))$ across checkpoints (smaller is better).}
\label{tab:mop_auc_results}
\scriptsize
\renewcommand{\arraystretch}{0.9}
\setlength{\tabcolsep}{3pt}
\begin{tabular}{|c|cc|ccc|ccc|ccc|ccc|}
\hline
 \multirow{2}{*}{Problem} & \multicolumn{2}{c|}{RDEx-MOP} & \multicolumn{3}{c|}{RDEx} & \multicolumn{3}{c|}{TEMOFNSGA3} & \multicolumn{3}{c|}{TFBCEIBEA} & \multicolumn{3}{c|}{TGFMMOEA} \\ \cline{2-15}
   & Mean & SD & Mean & SD & W & Mean & SD & W & Mean & SD & W & Mean & SD & W \\ \hline
 1 & 0.00 & 0.00 & \textbf{0.00} & \textbf{0.00} & = & 0.60 & 0.17 & + & 0.31 & 0.17 & + & 0.61 & 0.16 & + \\
 2 & 0.00 & 0.00 & \textbf{0.00} & \textbf{0.00} & = & 0.51 & 0.23 & + & 0.45 & 0.34 & + & 1.37 & 0.31 & + \\
 3 & \textbf{0.00} & \textbf{0.00} & 0.00 & 0.00 & = & 0.34 & 0.14 & + & 0.40 & 0.16 & + & 0.48 & 0.13 & + \\
 4 & 0.00 & 0.00 & \textbf{0.00} & \textbf{0.00} & = & 0.09 & 0.05 & + & 0.16 & 0.09 & + & 0.08 & 0.06 & + \\
 5 & 0.00 & 0.00 & \textbf{0.00} & \textbf{0.00} & = & 0.28 & 0.11 & + & 0.25 & 0.13 & + & 0.39 & 0.08 & + \\
 6 & \textbf{0.00} & \textbf{0.00} & 0.00 & 0.00 & = & 0.25 & 0.08 & + & 0.14 & 0.06 & + & 0.43 & 0.12 & + \\
 7 & 0.01 & 0.00 & \textbf{0.01} & \textbf{0.00} & - & 0.93 & 0.33 & + & 0.90 & 0.29 & + & 0.87 & 0.27 & + \\
 8 & \textbf{0.01} & \textbf{0.00} & 0.01 & 0.00 & = & 0.91 & 0.41 & + & 0.76 & 0.29 & + & 0.77 & 0.28 & + \\
 9 & \textbf{0.01} & \textbf{0.00} & 0.01 & 0.00 & = & 0.88 & 0.39 & + & 0.86 & 0.39 & + & 0.76 & 0.28 & + \\
 10 & 0.01 & 0.00 & \textbf{0.01} & \textbf{0.00} & = & 0.87 & 0.35 & + & 0.78 & 0.37 & + & 0.77 & 0.28 & + \\
\hline
 W/T/L & \multicolumn{2}{c|}{$-/-/-$} & \multicolumn{3}{c|}{0/9/1} & \multicolumn{3}{c|}{10/0/0} & \multicolumn{3}{c|}{10/0/0} & \multicolumn{3}{c|}{10/0/0} \\
\hline
\end{tabular}
\end{table}

\end{CJK}
\end{document}